\documentclass{article} 
\usepackage{iclr2024_conference,times}

\usepackage{amsmath,amsfonts,bm}

\def\eqref#1{equation~\ref{#1}}

\def\1{\bm{1}}

\DeclareMathAlphabet{\mathsfit}{\encodingdefault}{\sfdefault}{m}{sl}
\SetMathAlphabet{\mathsfit}{bold}{\encodingdefault}{\sfdefault}{bx}{n}

\usepackage{hyperref}
\usepackage{url}
\usepackage{graphicx}
\usepackage{caption}

\title{Black carbon plumes from gas flaring in North Africa identified from multi-spectral imagery with deep learning}

\author{Alexandre Tuel$^{*}$ \& Thomas Kerdreux\thanks{Equal contribution} \\
Galeio\\
Paris, 75014, France \\
\texttt{\{atuel,tkerdreux\}@galeio.fr} \\
\AND
Louis Thiry \\
Centre Inria Rennes - Bretagne Atlantique \\
Rennes, France \\
}

\iclrfinalcopy 
\begin{document}

\maketitle

\begin{abstract}
 Black carbon (BC) is an important pollutant aerosol emitted by numerous human activities, including gas flaring. Improper combustion in flaring activities can release large amounts of BC, which is harmful to human health and has a strong climate warming effect. To our knowledge, no study has ever directly monitored BC emissions from satellite imagery. Previous works quantified BC emissions indirectly, by applying emission coefficients to flaring volumes estimated from satellite imagery. Here, we develop a deep learning framework and apply it to Sentinel-2 imagery over North Africa during 2022 to detect and quantify BC emissions from gas flaring. We find that BC emissions in this region amount to about 1 million tCO$_{2,\mathrm{eq}}$, or 1 million passenger cars, more than a quarter of which are due to 10 sites alone. This work demonstrates the operational monitoring of BC emissions from flaring, a key step in implementing effective mitigation policies to reduce the climate impact of oil and gas operations.
\end{abstract}

\section{Introduction}

 Black carbon (BC), commonly known as soot, is a primary aerosol produced by incomplete combustion processes. It is made of almost pure carbon, and is an important atmospheric pollutant with significant health and environmental impacts \citep{Janssen2012,Bond2013,IPCCAR62021}. BC warms the Earth by absorbing visible solar radiation and reducing surface albedo when deposited onto snow or ice\citep{Ramanathan2008}. While BC is relatively short-lived in the atmosphere (4-10 days), it is estimated to account for about 0.1 W/m$^2$ of radiative forcing since pre-industrial times \citep{IPCCAR62021}. In addition, BC is an important constituent of fine particulate matter in air (PM$_{\mathrm{2.5}}$), and can easily combine with toxic chemicals \citep{EPA}. Reducing BC emissions can therefore both benefit human health and mitigate climate change.\\
 Major anthropogenic BC sources are fossil fuel and biomass combustion, with the largest being residential/commercial energy (heating and cooking) and industry \citep{Wang2014,Hoesly2018}. In particular, gas flaring is a significant contributor to industrial BC emissions. Flaring refers to the practice of burning excess natural gas to dispose of it during oil and gas extraction. An inefficient combustion process during flaring can cause vast amounts of BC to be emitted in large plumes \citep{EPA}. Flaring-related BC emissions are overall uncertain: average estimates range between 70 and 150 Gg/yr \citep{Caseiro2020,Huang2016}, but they could be as low as 20 Gg/yr and as high as 230 Gg/yr \citep{Faruolo2021} (100 Gg/yr amounts to about 1\% of global direct anthropogenic BC emissions \citep{Hoesly2018}).\\
 Such estimates are based on emission factors, which relate BC emissions to the volume or heating value of flared gas \citep{Faruolo2021}. Obtained from field measurements, emission factors for BC typically range from 0.1-1.5 g/m$^3$, and vary according to gas composition and flaring temperature \citep{McEwen2012,Caseiro2020,Boettcher2021}. Yet, the emission factor approach has several drawbacks which lead to large uncertainties in estimates. Flared gas volumes are uncertain, and emission factors themselves can vary significantly depending on local flare operating conditions, which are unknown and may vary over time \citep{Huang2016}. Few measurements are also available to estimate BC emission factors, and most are representative of well-operated flares with no visible BC plumes \citep{Pederstad2015,Allen2016,Klimont2017}.\\
 Here, we leverage satellite imagery to monitor BC emissions from gas flaring. Satellite data have already been extensively used to monitor flaring \citep[\textit{e.g.},][]{Elvidge2013,Caseiro2020}, specifically to locate gas flares and estimate flared gas volumes and flaring by-products \citep{Faruolo2021}. Satellite data are especially attractive as they can help tackle the legal and technical challenges of monitoring and reporting of gas flaring, as well as provide dense spatio-temporal flaring data \citep{Faruolo2021}. However, to our knowledge, all previous satellite-based attempts at quantifying BC emissions from flaring have relied on emission factors. We instead estimate BC emissions directly from satellite imagery, bypassing the need to characterise flares. We combine traditional flare-identification algorithms with a deep learning model to detect and quantify visible BC plumes around gas flares. We focus on North Africa, namely Algeria, Libya and Egypt, 3 of the 15 countries that flare the most gas according to recent estimates \citep{Caseiro2020}.

 \begin{figure}[h!]
    \centering
    \includegraphics[width=11cm]{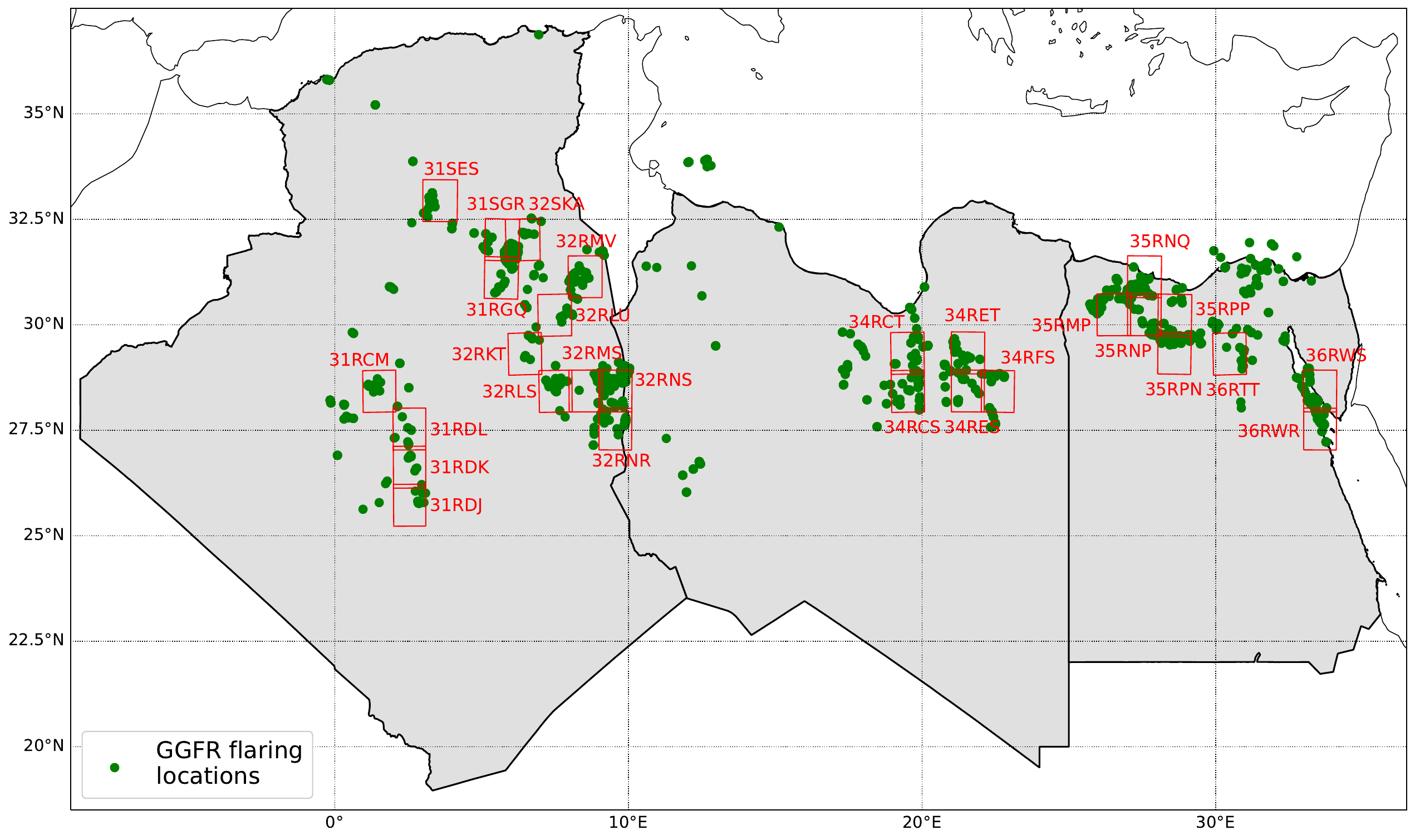}
    \caption{Map of our region of study (Algeria, Libya and Egypt) showing the GGFR flaring locations (green dots) and the set of 28 Sentinel-2 tiles we downloaded for this study (red squares, tile name shown in red).}
    \label{Fig_1}
 \end{figure}

\section{Proposed Methodology}

 We detect BC plumes in 2022 across Algeria, Libya and Egypt in top-of-atmosphere data from the European Space Agency's Sentinel-2 constellation \citep{Drusch2012}. This region concentrates a large number of gas flaring sites (Figure \ref{Fig_1}). We restrain ourselves to the 28 Sentinel-2 tiles ($\approx$ 340,000 km$^2$) which contain more than 5 flaring sites, based on the World Bank's Global Gas Flaring Reduction Partnership (GGFR)'s Global Flaring database \citep{WBdata}.\\

 \subsection{Model and training data}
 
 We develop a ConvLSTM model \cite{Shi2015} to detect BC plumes in satellite imagery. The model takes as input a time series of two RGB images, corresponding to two neighbouring satellite acquisitions, and returns a segmentation mask for BC plumes present in the second image. We train the models on 128$\times$128 randomly cropped RGB image time series of Sentinel-2 data (taken from a variety of tiles close to the ones of interest; Figure \ref{Fig_1}) in which we embed synthetic BC plumes. These synthetic plumes are simulated with a custom 2D LES plume model which implements the 2D Euler equations with finite volume and uses an implicit (\textit{i.e.} numerical) subgrid scale term \citep{Grinstein2007}. We use an FFT-based pseudo-spectral solver to solve the Poisson equation. Our implicit closure lies in the advection scheme: we use WENO-3 reconstruction \citep{Jiang1996} to compute fluxes. This yields fast synthetic plumes simulations, allowing us to generate a large synthetic plumes set in a reasonable computational time.\\
 The reason we use the three RGB channels only (out of the 13 multispectral bands of Sentinel-2) is that the absorption spectrum of BC outside the RGB domain is not well known. Restricting ourselves to RGB thus allows us to easily integrate the synthetic plumes to Sentinel-2 RGB scenes by dimming the original RGB scene brightness by a factor proportional to the plume density (from 0 (no plume) to 1 (total extinction due to plume)) (Figure \ref{Fig_S2}). However, we do use the other 10 channels in the post-processing to remove false positives (see below).\\

 \subsection{Formatting of Sentinel-2 data}
 Because flares are not continuously lit, or can be masked by clouds, we first detect flares for each tile and each acquisition with the tri-band spectral algorithms developed for Sentinel-2: the Normalized Hotspot Indices \citep{Marchese2019} (NHI) and the Thermal Anomaly Index \citep{Liu2021} (TAI). The performance and limitation of such algorithms have been thoroughly studied \citep{Genzano2020,Sharma2017,Fisher2019}. We have found them to be very reliable, with few false negatives, though bright, low-level clouds are sometimes mistaken for lit flares. We then group together flares separated by less than 1 km (for which it is difficult to separately attribute BC plumes). Finally, we crop Sentinel-2 scenes around each flare cluster, by embedding the cluster in a square and expanding this square by 256 pixels in each direction (the crop window is therefore at least $512\times512$, or $\approx$ $5\times5$ km).

 \subsection{Post-processing of model output}
 We first average the predictions from 10 separate models to retain plumes detected by at least half of the models. We then filter the detections to remove potential false positives by using a lightGBM classifier that takes as input the 13 spectral channels of Sentinel-2 data. The classifier has false positive and negative rates of about 2-3\% each. We subsequently group together all predictions for a given tile and date. From this average, we identify each individual BC plume which we assign to the nearest cluster of flares. Finally, we estimate BC emission rates using the Integrated Mass Enhancement (IME) method \citep{Frankenberg2016,Varon2018}, as described in Appendix \ref{appendix}.

\section{Evaluation}

 We identify a total of 1963 individual flares, active for at least one acquisition date during 2022 (Figure \ref{Fig_S1}). They are grouped within 1104 clusters of flares, the vast majority (83\%) of which are active for 10\% or less of the time. About half of flare clusters are located in Algeria (51\%), 12\% in Libya and 37\% in Egypt. Algeria accounts for 55\% of total flaring activity, Libya for 21\% and Egypt for 24\%.

\begin{figure}[h!]
    \centering
    \includegraphics[width=11cm]{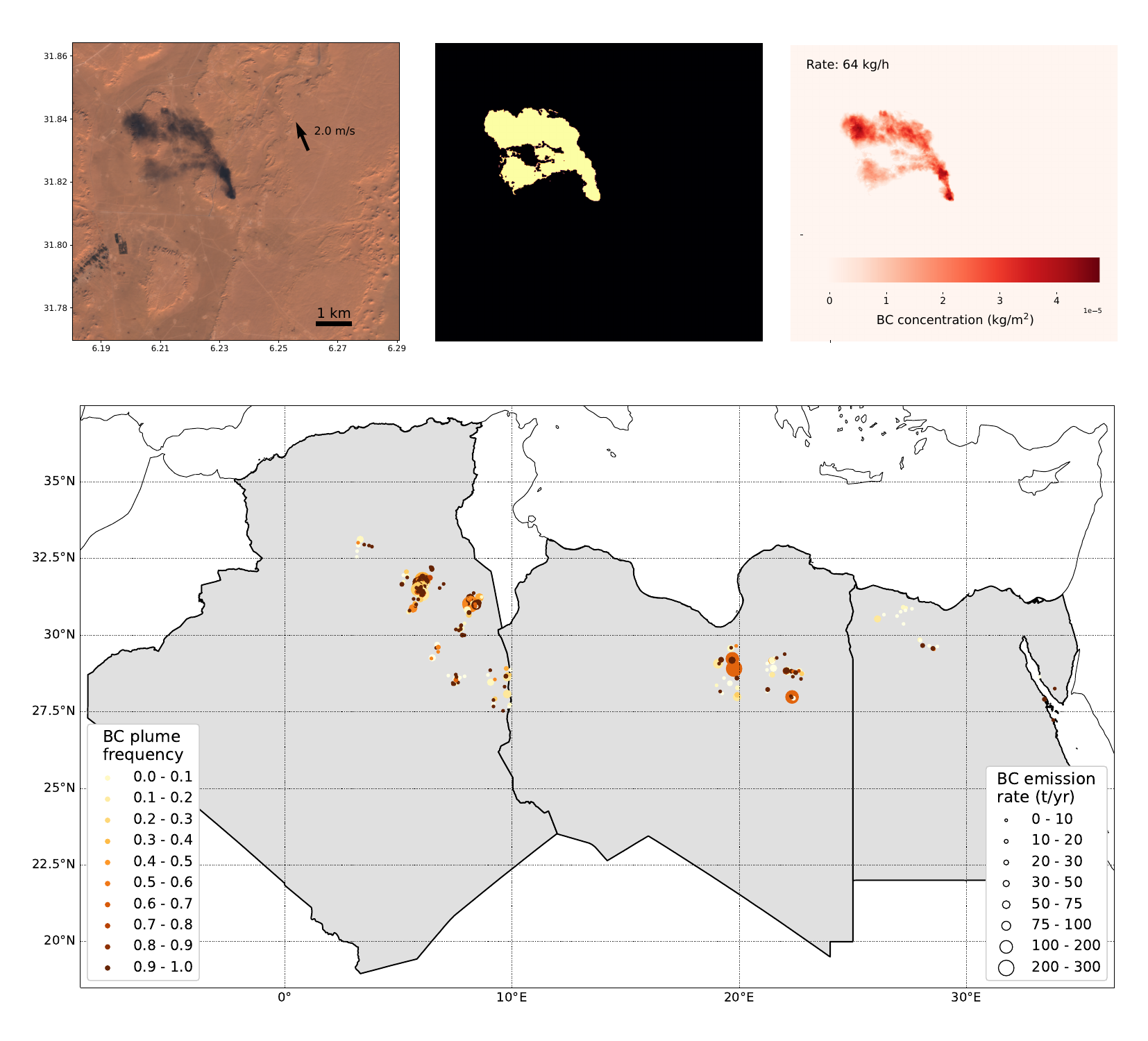}
    \caption{Top panels: example of detected BC plume in Algeria; from left to right: RGB image, plume prediction mask, and BC concentration map (obtained with $\mu=8$ m$^2$/g, see Appendix \ref{appendix}). Bottom panel: Map of BC emission rates (in t/yr) across North Africa. The size and colour of the dots depends on the annual flaring frequency of each location.}
    \label{Fig_3}
 \end{figure}

 BC plumes are overall well-captured by our deep learning model (Figure \ref{Fig_3}). All large plumes are correctly identified by the algorithm, but very small plumes ($\leq$ 500 m in maximum extent) are often missed. Small and medium plumes may also be missed when atmospheric non-BC aerosol concentration (\textit{e.g.}, sand particles) is high and causes haze, which frequently happens in the desert. The model also has a slight tendency to underestimate the true extent of the plumes, especially when these are very diluted.\\
 We detect BC plumes at 372 clusters of flares (297 in Algeria, 55 in Libya and 20 in Egypt). BC plume frequency is highest in Algeria (BC plumes detected in 65\% of active flares), followed by Libya (49\%) and Egypt (34\% only). Annual BC emission rates range between 0.01 and 270 t/yr, with 65\% of locations emitting less than 10 t/yr (Figure \ref{Fig_S3}). In total, we estimate that the flares we monitored emitted 6.2 Gg of BC in 2022 [3.1-12.5 Gg]. Algeria accounts for 75\% of this total (4.7 Gg [2.3-9.3 Gg]), Libya for 23\% (1.4 Gg [0.7-2.8 Gg]) and Egypt for 2\% (0.14 Gg [0.07-0.29 Gg]). Regional emissions are dominated by a very small number of locations: the 10 clusters emitting more than 100 t/yr (7 in Algeria, 3 in Libya.) are responsible for 27\% of total emissions (Figure \ref{Fig_S4}).
 
\section{Conclusion}

 This work demonstrates the possibility to monitor flaring BC emissions at large scale over desert regions and in real time. BC plumes are a frequent feature of gas flaring across North Africa, especially in Algeria and Libya, some of which even stretch over several square kilometers (Figure \ref{Fig_3}). Interestingly, BC emissions widely differ across flares. Some flares appear particularly inefficient and emit large quantities of BC, while some very active ones have much smaller BC emissions. BC plumes also tend to be intermittent. For most flares that are active $>$ 20\% of the time, BC plumes are detected in only 10-20\% of acquisitions (Figure \ref{Fig_S3}). This suggests that flare operating conditions vary widely during the year.\\
 There remain large uncertainties in our BC emission estimates. The sources of this uncertainty include sampling uncertainties (due to infrequent satellite coverage or cloud cover), model uncertainties (accuracy and estimation of plume masks), and uncertainties in plume rate estimates (IME method, imperfect wind data). Still, our estimates are comparable to previous ones based on emission factors \citep[\textit{e.g.},][]{Caseiro2020}. The originality of our method consists in directly detecting BC emissions and being able to pinpoint the most emissive sites.\\
 BC being an important -- but short-lived -- climate forcer, reducing flaring-related BC emissions where flaring infrastructure is particularly inefficient can yield substantial benefits at the short term. Assuming a 100-year global warming potential of BC of 500-1000 \citep{Reddy2007,Bond2013}, flaring BC emissions in North Africa in 2022 are equivalent to 3.1-6.2 million tCO$_2$, or 0.7-1.3 million gas-powered passenger cars \citep{EPA2023}. Detected BC plumes also indicate improper flare combustion, which may also release other pollutants, like methane, that have even larger environmental impacts. Most of this is accounted for by a handful of sites, meaning that limited investments in efficient flaring infrastructure or to improve operating practices are realistic and could considerably decrease the environmental and health impacts of gas flaring in North Africa.

\newpage

\bibliography{references_workshop}

\begin{thebibliography}{38}
\providecommand{\natexlab}[1]{#1}
\providecommand{\url}[1]{\texttt{#1}}
\expandafter\ifx\csname urlstyle\endcsname\relax
  \providecommand{\doi}[1]{doi: #1}\else
  \providecommand{\doi}{doi: \begingroup \urlstyle{rm}\Url}\fi

\bibitem[Allen et~al.(2016)Allen, Smith, Torres, and Saldaña]{Allen2016}
David~T Allen, Denzil Smith, Vincent~M Torres, and Felipe~Cardoso Saldaña.
\newblock {Carbon dioxide, methane and black carbon emissions from upstream oil and gas flaring in the United States}.
\newblock \emph{Current Opinion in Chemical Engineering}, 13:\penalty0 119--123, 2016.
\newblock ISSN 2211-3398.
\newblock \doi{10.1016/j.coche.2016.08.014}.
\newblock URL \url{https://www.sciencedirect.com/science/article/pii/S2211339816300569}.
\newblock Energy and Environmental Engineering / Reaction engineering and catalysis.

\bibitem[Bond et~al.(2013)Bond, Doherty, Fahey, Forster, Berntsen, DeAngelo, Flanner, Ghan, Kärcher, Koch, Kinne, Kondo, Quinn, Sarofim, Schultz, Schulz, Venkataraman, Zhang, Zhang, Bellouin, Guttikunda, Hopke, Jacobson, Kaiser, Klimont, Lohmann, Schwarz, Shindell, Storelvmo, Warren, and Zender]{Bond2013}
T.~C. Bond, S.~J. Doherty, D.~W. Fahey, P.~M. Forster, T.~Berntsen, B.~J. DeAngelo, M.~G. Flanner, S.~Ghan, B.~Kärcher, D.~Koch, S.~Kinne, Y.~Kondo, P.~K. Quinn, M.~C. Sarofim, M.~G. Schultz, M.~Schulz, C.~Venkataraman, H.~Zhang, S.~Zhang, N.~Bellouin, S.~K. Guttikunda, P.~K. Hopke, M.~Z. Jacobson, J.~W. Kaiser, Z.~Klimont, U.~Lohmann, J.~P. Schwarz, D.~Shindell, T.~Storelvmo, S.~G. Warren, and C.~S. Zender.
\newblock {Bounding the role of black carbon in the climate system: A scientific assessment}.
\newblock \emph{Journal of Geophysical Research: Atmospheres}, 118\penalty0 (11):\penalty0 5380--5552, 2013.
\newblock \doi{10.1002/jgrd.50171}.
\newblock URL \url{https://agupubs.onlinelibrary.wiley.com/doi/abs/10.1002/jgrd.50171}.

\bibitem[Böttcher et~al.(2021)Böttcher, Paunu, Kupiainen, Zhizhin, Matveev, Savolahti, Klimont, Väätäinen, Lamberg, and Karvosenoja]{Boettcher2021}
Kristin Böttcher, Ville-Veikko Paunu, Kaarle Kupiainen, Mikhail Zhizhin, Alexey Matveev, Mikko Savolahti, Zbigniew Klimont, Sampsa Väätäinen, Heikki Lamberg, and Niko Karvosenoja.
\newblock {Black carbon emissions from flaring in Russia in the period 2012–2017}.
\newblock \emph{Atmospheric Environment}, 254:\penalty0 118390, 2021.
\newblock ISSN 1352-2310.
\newblock \doi{10.1016/j.atmosenv.2021.118390}.
\newblock URL \url{https://www.sciencedirect.com/science/article/pii/S1352231021002090}.

\bibitem[Caseiro et~al.(2020)Caseiro, Gehrke, R\"ucker, Leimbach, and Kaiser]{Caseiro2020}
A.~Caseiro, B.~Gehrke, G.~R\"ucker, D.~Leimbach, and J.~W. Kaiser.
\newblock {Gas flaring activity and black carbon emissions in 2017 derived from the Sentinel-3A Sea and Land Surface Temperature Radiometer}.
\newblock \emph{Earth System Science Data}, 12\penalty0 (3):\penalty0 2137--2155, 2020.
\newblock \doi{10.5194/essd-12-2137-2020}.
\newblock URL \url{https://essd.copernicus.org/articles/12/2137/2020/}.

\bibitem[Chung et~al.(2012)Chung, Lee, and Müller]{Chung2012}
ChulE. Chung, Kyunghwa Lee, and Detlef Müller.
\newblock Effect of internal mixture on black carbon radiative forcing.
\newblock \emph{Tellus B: Chemical and Physical Meteorology}, 64\penalty0 (1):\penalty0 10925, 2012.
\newblock \doi{10.3402/tellusb.v64i0.10925}.
\newblock URL \url{https://doi.org/10.3402/tellusb.v64i0.10925}.

\bibitem[Conrad \& Johnson(2019)Conrad and Johnson]{Conrad2019}
B.M. Conrad and M.R. Johnson.
\newblock {Mass absorption cross-section of flare-generated black carbon: Variability, predictive model, and implications}.
\newblock \emph{Carbon}, 149:\penalty0 760--771, 2019.
\newblock ISSN 0008-6223.
\newblock \doi{10.1016/j.carbon.2019.04.086}.
\newblock URL \url{https://www.sciencedirect.com/science/article/pii/S0008622319304269}.

\bibitem[Drusch et~al.(2012)Drusch, {Del Bello}, Carlier, Colin, Fernandez, Gascon, Hoersch, Isola, Laberinti, Martimort, Meygret, Spoto, Sy, Marchese, and Bargellini]{Drusch2012}
M.~Drusch, U.~{Del Bello}, S.~Carlier, O.~Colin, V.~Fernandez, F.~Gascon, B.~Hoersch, C.~Isola, P.~Laberinti, P.~Martimort, A.~Meygret, F.~Spoto, O.~Sy, F.~Marchese, and P.~Bargellini.
\newblock {Sentinel-2: ESA's Optical High-Resolution Mission for GMES Operational Services}.
\newblock \emph{Remote Sensing of Environment}, 120:\penalty0 25--36, 2012.
\newblock ISSN 0034-4257.
\newblock \doi{10.1016/j.rse.2011.11.026}.
\newblock URL \url{https://www.sciencedirect.com/science/article/pii/S0034425712000636}.
\newblock The Sentinel Missions - New Opportunities for Science.

\bibitem[Elvidge et~al.(2013)Elvidge, Zhizhin, Hsu, and Baugh]{Elvidge2013}
Christopher~D. Elvidge, Mikhail Zhizhin, Feng-Chi Hsu, and Kimberly~E. Baugh.
\newblock {VIIRS Nightfire: Satellite Pyrometry at Night}.
\newblock \emph{Remote Sensing}, 5\penalty0 (9):\penalty0 4423--4449, 2013.
\newblock ISSN 2072-4292.
\newblock \doi{10.3390/rs5094423}.
\newblock URL \url{https://www.mdpi.com/2072-4292/5/9/4423}.

\bibitem[EPA(2012)]{EPA}
EPA.
\newblock Report to congress on black carbon.
\newblock Technical report, United States Environmental Protection Agency, 2012.
\newblock URL \url{https://19january2017snapshot.epa.gov/www3/airquality/blackcarbon/2012report/fullreport.pdf}.
\newblock last accessed August 12, 2023.

\bibitem[EPA(2023)]{EPA2023}
EPA.
\newblock Tailpipe greenhouse gas emissions from a typical passenger vehicle, 2023.
\newblock URL \url{https://nepis.epa.gov/Exe/ZyPDF.cgi?Dockey=P1017FP5.pdf}.
\newblock last accessed August 16, 2023.

\bibitem[Faruolo et~al.(2021)Faruolo, Caseiro, Lacava, and Kaiser]{Faruolo2021}
Mariapia Faruolo, Alexandre Caseiro, Teodosio Lacava, and Johannes~W. Kaiser.
\newblock {Gas Flaring: A Review Focused On Its Analysis From Space}.
\newblock \emph{IEEE Geoscience and Remote Sensing Magazine}, 9\penalty0 (1):\penalty0 258--281, 2021.
\newblock \doi{10.1109/MGRS.2020.3007232}.

\bibitem[Fierce et~al.(2020)Fierce, Onasch, Cappa, Mazzoleni, China, Bhandari, Davidovits, Fischer, Helgestad, Lambe, Sedlacek, Smith, and Wolff]{Fierce2020}
Laura Fierce, Timothy~B. Onasch, Christopher~D. Cappa, Claudio Mazzoleni, Swarup China, Janarjan Bhandari, Paul Davidovits, D.~Al Fischer, Taylor Helgestad, Andrew~T. Lambe, Arthur~J. Sedlacek, Geoffrey~D. Smith, and Lindsay Wolff.
\newblock Radiative absorption enhancements by black carbon controlled by particle-to-particle heterogeneity in composition.
\newblock \emph{Proceedings of the National Academy of Sciences}, 117\penalty0 (10):\penalty0 5196--5203, 2020.
\newblock \doi{10.1073/pnas.1919723117}.
\newblock URL \url{https://www.pnas.org/doi/abs/10.1073/pnas.1919723117}.

\bibitem[Fisher \& Wooster(2019)Fisher and Wooster]{Fisher2019}
Daniel Fisher and Martin~J Wooster.
\newblock {Multi-decade global gas flaring change inventoried using the ATSR-1, ATSR-2, AATSR and SLSTR data records}.
\newblock \emph{Remote Sensing of Environment}, 232:\penalty0 111298, 2019.

\bibitem[Frankenberg et~al.(2016)Frankenberg, Thorpe, Thompson, Hulley, Kort, Vance, Borchardt, Krings, Gerilowski, Sweeney, Conley, Bue, Aubrey, Hook, and Green]{Frankenberg2016}
Christian Frankenberg, Andrew~K. Thorpe, David~R. Thompson, Glynn Hulley, Eric~Adam Kort, Nick Vance, Jakob Borchardt, Thomas Krings, Konstantin Gerilowski, Colm Sweeney, Stephen Conley, Brian~D. Bue, Andrew~D. Aubrey, Simon Hook, and Robert~O. Green.
\newblock {Airborne methane remote measurements reveal heavy-tail flux distribution in Four Corners region}.
\newblock \emph{Proceedings of the National Academy of Sciences}, 113\penalty0 (35):\penalty0 9734--9739, 2016.
\newblock \doi{10.1073/pnas.1605617113}.
\newblock URL \url{https://www.pnas.org/doi/abs/10.1073/pnas.1605617113}.

\bibitem[Genzano et~al.(2020)Genzano, Pergola, and Marchese]{Genzano2020}
Nicola Genzano, Nicola Pergola, and Francesco Marchese.
\newblock {A Google Earth Engine tool to investigate, map and monitor volcanic thermal anomalies at global scale by means of mid-high spatial resolution satellite data}.
\newblock \emph{Remote Sensing}, 12\penalty0 (19):\penalty0 3232, 2020.

\bibitem[Grinstein et~al.(2007)Grinstein, Margolin, and Rider]{Grinstein2007}
Fernando~F Grinstein, Len~G Margolin, and William~J Rider.
\newblock \emph{Implicit large eddy simulation}, volume~10.
\newblock Cambridge university press Cambridge, 2007.

\bibitem[Hoesly et~al.(2018)Hoesly, Smith, Feng, Klimont, Janssens-Maenhout, Pitkanen, Seibert, Vu, Andres, Bolt, Bond, Dawidowski, Kholod, Kurokawa, Li, Liu, Lu, Moura, O'Rourke, and Zhang]{Hoesly2018}
R.~M. Hoesly, S.~J. Smith, L.~Feng, Z.~Klimont, G.~Janssens-Maenhout, T.~Pitkanen, J.~J. Seibert, L.~Vu, R.~J. Andres, R.~M. Bolt, T.~C. Bond, L.~Dawidowski, N.~Kholod, J.-I. Kurokawa, M.~Li, L.~Liu, Z.~Lu, M.~C.~P. Moura, P.~R. O'Rourke, and Q.~Zhang.
\newblock {Historical (1750--2014) anthropogenic emissions of reactive gases and aerosols from the Community Emissions Data System (CEDS)}.
\newblock \emph{Geoscientific Model Development}, 11\penalty0 (1):\penalty0 369--408, 2018.
\newblock \doi{10.5194/gmd-11-369-2018}.
\newblock URL \url{https://gmd.copernicus.org/articles/11/369/2018/}.

\bibitem[Huang \& Fu(2016)Huang and Fu]{Huang2016}
Kan Huang and Joshua~S. Fu.
\newblock {A global gas flaring black carbon emission rate dataset from 1994 to 2012}.
\newblock \emph{Scientific Data}, 3\penalty0 (1):\penalty0 160104, 2016.
\newblock ISSN 2052-4463.
\newblock \doi{10.1038/sdata.2016.104}.
\newblock URL \url{https://doi.org/10.1038/sdata.2016.104}.

\bibitem[Janssen et~al.(2012)Janssen, Gerlofs-Nijland, Lanki, Salonen, Cassee, Hoek, Fischer, Brunekreef, and Krzyzanowski]{Janssen2012}
Nicole~AH Janssen, Miriam~E Gerlofs-Nijland, Timo Lanki, Raimo~O Salonen, Flemming Cassee, Gerard Hoek, Paul Fischer, Bert Brunekreef, and Michal Krzyzanowski.
\newblock Health effects of black carbon.
\newblock Technical report, World Health Organization, Regional Office for Europe, 2012.
\newblock URL \url{https://apps.who.int/iris/handle/10665/352615}.

\bibitem[Jiang \& Shu(1996)Jiang and Shu]{Jiang1996}
Guang-Shan Jiang and Chi-Wang Shu.
\newblock {Efficient implementation of weighted ENO schemes}.
\newblock \emph{Journal of computational physics}, 126\penalty0 (1):\penalty0 202--228, 1996.

\bibitem[Kelesidis et~al.(2021)Kelesidis, Bruun, and Pratsinis]{Kelesidis2021}
Georgios~A. Kelesidis, Christian~A. Bruun, and Sotiris~E. Pratsinis.
\newblock The impact of organic carbon on soot light absorption.
\newblock \emph{Carbon}, 172:\penalty0 742--749, 2021.
\newblock ISSN 0008-6223.
\newblock \doi{10.1016/j.carbon.2020.10.032}.
\newblock URL \url{https://www.sciencedirect.com/science/article/pii/S0008622320309969}.

\bibitem[Kirchstetter et~al.(2004)Kirchstetter, Novakov, and Hobbs]{Kirchstetter2004}
Thomas~W. Kirchstetter, T.~Novakov, and Peter~V. Hobbs.
\newblock Evidence that the spectral dependence of light absorption by aerosols is affected by organic carbon.
\newblock \emph{Journal of Geophysical Research: Atmospheres}, 109\penalty0 (D21), 2004.
\newblock \doi{10.1029/2004JD004999}.
\newblock URL \url{https://agupubs.onlinelibrary.wiley.com/doi/abs/10.1029/2004JD004999}.

\bibitem[Klimont et~al.(2017)Klimont, Kupiainen, Heyes, Purohit, Cofala, Rafaj, Borken-Kleefeld, and Sch\"opp]{Klimont2017}
Z.~Klimont, K.~Kupiainen, C.~Heyes, P.~Purohit, J.~Cofala, P.~Rafaj, J.~Borken-Kleefeld, and W.~Sch\"opp.
\newblock Global anthropogenic emissions of particulate matter including black carbon.
\newblock \emph{Atmospheric Chemistry and Physics}, 17\penalty0 (14):\penalty0 8681--8723, 2017.
\newblock \doi{10.5194/acp-17-8681-2017}.
\newblock URL \url{https://acp.copernicus.org/articles/17/8681/2017/}.

\bibitem[Lee et~al.(2022)Lee, Gorkowski, Meyer, Benedict, Aiken, and Dubey]{Lee2022}
James~E. Lee, Kyle Gorkowski, Aaron~G. Meyer, Katherine~B. Benedict, Allison~C. Aiken, and Manvendra~K. Dubey.
\newblock {Wildfire Smoke Demonstrates Significant and Predictable Black Carbon Light Absorption Enhancements}.
\newblock \emph{Geophysical Research Letters}, 49\penalty0 (14):\penalty0 e2022GL099334, 2022.
\newblock \doi{10.1029/2022GL099334}.
\newblock URL \url{https://agupubs.onlinelibrary.wiley.com/doi/abs/10.1029/2022GL099334}.

\bibitem[Liu et~al.(2021)Liu, Zhi, Xu, Xu, and Wu]{Liu2021}
Yongxue Liu, Weifeng Zhi, Bihua Xu, Wenxuan Xu, and Wei Wu.
\newblock {Detecting high-temperature anomalies from Sentinel-2 MSI images}.
\newblock \emph{ISPRS Journal of Photogrammetry and Remote Sensing}, 177:\penalty0 174--193, 2021.

\bibitem[Marchese et~al.(2019)Marchese, Genzano, Neri, Falconieri, Mazzeo, and Pergola]{Marchese2019}
Francesco Marchese, Nicola Genzano, Marco Neri, Alfredo Falconieri, Giuseppe Mazzeo, and Nicola Pergola.
\newblock {A multi-channel algorithm for mapping volcanic thermal anomalies by means of Sentinel-2 MSI and Landsat-8 OLI data}.
\newblock \emph{Remote Sensing}, 11\penalty0 (23):\penalty0 2876, 2019.

\bibitem[McEwen \& Johnson(2012)McEwen and Johnson]{McEwen2012}
James~D.N. McEwen and Matthew~R. Johnson.
\newblock Black carbon particulate matter emission factors for buoyancy-driven associated gas flares.
\newblock \emph{Journal of the Air \& Waste Management Association}, 62\penalty0 (3):\penalty0 307--321, 2012.
\newblock \doi{10.1080/10473289.2011.650040}.
\newblock URL \url{https://doi.org/10.1080/10473289.2011.650040}.

\bibitem[Molod et~al.(2015)Molod, Takacs, Suarez, and Bacmeister]{Molod2015}
A.~Molod, L.~Takacs, M.~Suarez, and J.~Bacmeister.
\newblock {Development of the GEOS-5 atmospheric general circulation model: evolution from MERRA to MERRA2}.
\newblock \emph{Geoscientific Model Development}, 8\penalty0 (5):\penalty0 1339--1356, 2015.
\newblock \doi{10.5194/gmd-8-1339-2015}.
\newblock URL \url{https://gmd.copernicus.org/articles/8/1339/2015/}.

\bibitem[{NOAA, the Payne Institute at the Colorado School of Mines, World Bank/GGFR}(2023)]{WBdata}
{NOAA, the Payne Institute at the Colorado School of Mines, World Bank/GGFR}.
\newblock Flare gas volumes, 2023.
\newblock URL \url{https://thedocs.worldbank.org/en/doc/1ccd33e2f13ebc7cfd9647fdebc843f2-0400072023/related/GGFR-Flaring-Dashboard-Data-March292023.xlsx}.

\bibitem[Pederstad et~al.(2015)Pederstad, D., R., Saunier, and Holm]{Pederstad2015}
A.~Pederstad, Smith~J. D., Jackson R., S.~Saunier, and T.~Holm.
\newblock Assessment of flare strategies, techniques for reduction of flaring and associated emissions, emission factors and methods for determination of emissions to air from flaring.
\newblock Technical report, Carbon Limits, 2015.
\newblock URL \url{https://www.miljodirektoratet.no/globalassets/publikasjoner/M312/M312.pdf}.
\newblock last accessed August 12, 2023.

\bibitem[Ramanathan \& Carmichael(2008)Ramanathan and Carmichael]{Ramanathan2008}
V.~Ramanathan and G.~Carmichael.
\newblock {Global and regional climate changes due to black carbon}.
\newblock \emph{Nature Geoscience}, 1\penalty0 (4):\penalty0 221--227, 2008.
\newblock \doi{10.1038/ngeo156}.
\newblock URL \url{https://doi.org/10.1038/ngeo156}.

\bibitem[Reddy \& Boucher(2007)Reddy and Boucher]{Reddy2007}
M.~Shekar Reddy and Olivier Boucher.
\newblock Climate impact of black carbon emitted from energy consumption in the world's regions.
\newblock \emph{Geophysical Research Letters}, 34\penalty0 (11), 2007.
\newblock \doi{10.1029/2006GL028904}.
\newblock URL \url{https://agupubs.onlinelibrary.wiley.com/doi/abs/10.1029/2006GL028904}.

\bibitem[Sharma et~al.(2017)Sharma, Wang, and Lennartson]{Sharma2017}
Ambrish Sharma, Jun Wang, and Elizabeth~M Lennartson.
\newblock Intercomparison of modis and viirs fire products in khanty-mansiysk russia: Implications for characterizing gas flaring from space.
\newblock \emph{Atmosphere}, 8\penalty0 (6):\penalty0 95, 2017.

\bibitem[Shi et~al.(2015)Shi, Chen, Wang, Yeung, Wong, and Woo]{Shi2015}
Xingjian Shi, Zhourong Chen, Hao Wang, Dit-Yan Yeung, Wai-Kin Wong, and Wang-chun Woo.
\newblock {Convolutional LSTM network: A machine learning approach for precipitation nowcasting}.
\newblock \emph{Advances in neural information processing systems}, 28, 2015.

\bibitem[Szopa et~al.(2021)Szopa, Naik, Adhikary, Artaxo, Berntsen, Collins, Fuzzi, Gallardo, Kiendler-Scharr, Klimont, Liao, Unger, and Zanis]{IPCCAR62021}
S.~Szopa, V.~Naik, B.~Adhikary, P.~Artaxo, T.~Berntsen, W.D. Collins, S.~Fuzzi, L.~Gallardo, A.~Kiendler-Scharr, Z.~Klimont, H.~Liao, N.~Unger, and P.~Zanis.
\newblock {Short-Lived Climate Forcers}.
\newblock In \emph{Climate Change 2021: The Physical Science Basis. Contribution of Working Group I to the Sixth Assessment Report of the Intergovernmental Panel on Climate Change [Masson-Delmotte, V., P. Zhai, A. Pirani, S.L. Connors, C. Péan, S. Berger, N. Caud, Y. Chen, L. Goldfarb, M.I. Gomis, M. Huang, K. Leitzell, E. Lonnoy, J.B.R. Matthews, T.K. Maycock, T. Waterfield, O. Yelekçi, R. Yu, and B. Zhou (eds.)].}, pp.\  817--922. Cambridge University Press, 2021.
\newblock \doi{10.1017/9781009157896.008}.
\newblock URL \url{https://doi.org/10.1017/9781009157896.008}.

\bibitem[Varon et~al.(2018)Varon, Jacob, McKeever, Jervis, Durak, Xia, and Huang]{Varon2018}
Daniel~J Varon, Daniel~J Jacob, Jason McKeever, Dylan Jervis, Berke~OA Durak, Yan Xia, and Yi~Huang.
\newblock Quantifying methane point sources from fine-scale satellite observations of atmospheric methane plumes.
\newblock \emph{Atmospheric Measurement Techniques}, 11\penalty0 (10):\penalty0 5673--5686, 2018.
\newblock \doi{10.5194/amt-11-5673-2018}.

\bibitem[Wang et~al.(2014)Wang, Tao, Shen, Huang, Chen, Balkanski, Boucher, Ciais, Shen, Li, Zhang, Chen, Lin, Su, Li, Liu, and Liu]{Wang2014}
Rong Wang, Shu Tao, Huizhong Shen, Ye~Huang, Han Chen, Yves Balkanski, Olivier Boucher, Philippe Ciais, Guofeng Shen, Wei Li, Yanyan Zhang, Yuanchen Chen, Nan Lin, Shu Su, Bengang Li, Junfeng Liu, and Wenxin Liu.
\newblock {Trend in Global Black Carbon Emissions from 1960 to 2007}.
\newblock \emph{Environmental Science \& Technology}, 48\penalty0 (12):\penalty0 6780--6787, 2014.
\newblock \doi{10.1021/es5021422}.
\newblock URL \url{https://doi.org/10.1021/es5021422}.

\bibitem[Weyant et~al.(2016)Weyant, Shepson, Subramanian, Cambaliza, Heimburger, McCabe, Baum, Stirm, and Bond]{Weyant2016}
Cheryl~L. Weyant, Paul~B. Shepson, R.~Subramanian, Maria O.~L. Cambaliza, Alexie Heimburger, David McCabe, Ellen Baum, Brian~H. Stirm, and Tami~C. Bond.
\newblock {Black Carbon Emissions from Associated Natural Gas Flaring}.
\newblock \emph{Environmental Science \& Technology}, 50\penalty0 (4):\penalty0 2075--2081, 2016.
\newblock \doi{10.1021/acs.est.5b04712}.
\newblock URL \url{https://doi.org/10.1021/acs.est.5b04712}.

\end{thebibliography}
\bibliographystyle{iclr2024_conference}

\newpage

\appendix
\section{Appendix}
\subsection{Estimating BC emission rates}\label{appendix}

 We give here a brief description of the Integrated Mass Enhancement (IME) method \citep{Frankenberg2016,Varon2018} we use to quantify BC plume emission rates. Denoting by $\Delta\Omega$ the BC concentration field (in kg m$^{-2}$) and $\mathcal{P}$ the plume mask, the total BC mass within the plume is equal to:
 \begin{equation}
     IME = \sum_{i\in \mathcal{P}} \Delta\Omega(i) A(i)
 \end{equation}
 \noindent where $A(i)$ is the area of grid cell $i$ (in m$^2$). The plume rate $Q$ is then empirically related to $IME$ by way of an effective wind speed $U_{eff}$ (in m s$^{-1}$), and an effective plume size $L$ (in m) through:
 \begin{equation}
     Q = \frac{U_{eff} IME}{L}
 \end{equation}
 \citet{Varon2018} found in the case of methane that $U_{eff}$ could be derived from the local 10-meter wind speed $U_{10}$ through $U_{eff} = 0.9\log(U_{10})+0.6$. We obtain 10-meter wind speed from the hourly GEOS-FP global reanalysis \citep{Molod2015}. Finally, we retrieve $\Delta\Omega$ with radiative transfer theory and the Beer-Lambert relationship. The latter relates light extinction in the Sentinel-2 B3 band (green; $\approx$535-585 nm) to total BC column concentration as follows:
 \begin{equation}
     \frac{I^{B3}_{BC}}{I^{B3}_0} = \int_{\lambda\in B3}w(\lambda)e^{-\gamma \mu \rho}d\lambda
     \label{eq:BL}
 \end{equation}
 \noindent where $\gamma$ is the atmospheric mass factor (function of the solar zenith and satellite viewing angle), $\mu$ is the BC mass extinction coefficient (in m$^2$/g) and $\rho$ the total BC column concentration (in g/m$^2$). $\frac{I^{B3}_{BC}}{I^{B3}_0}$ is the B3 transmittance between the target scene (which includes a BC plume) and a reference scene (in which the BC plume is absent). $w(\lambda)$ is the B3 spectral weight. In theory, $\mu$ is not constant with $\lambda$, but the difference is small over the limited wavelength range of B3. We use experimental values of $\mu$ at 550 nm given in the literature. Estimates vary across studies, but most fall in the 4-20 m$^2$/g range \citep{Weyant2016,Conrad2019,Lee2022}.\\
 We assume in equation (\ref{eq:BL}) that BC is solely responsible for light extinction in B3. In particular, we neglect the effect of organic carbon (OC), also present in the plume \citep[up to 20\% of the flared aerosol mass;][]{McEwen2012,Klimont2017}. The mass extinction coefficient of OC is however much smaller than that of BC, in particular around 550 nm \citep{Kirchstetter2004}.\\
 $\mu$ is also known to depend on its molecular structure and OC content \citep{Chung2012,Fierce2020}. OC coating of BC aerosols indeed strongly modifies its radiative properties, namely decreasing its absorption efficiency \citep{Kelesidis2021}. To take these various uncertainties into account, we estimate BC plume emissions using a median value of $\mu=8$ m$^2$/g and extreme values of $\mu=4$ and $\mu=20$ m$^2$/g.\\ 
 Finally, we translate the series of BC emission rates for each plume (in t/h) into annual emission rates (in t/yr) as follows:
 \begin{equation}
     Q_{t/yr} = \overline{Q_{t/h}} \times f_{\mathrm{flare}} \times f_{\mathrm{plume}} \times 365 \times 24
 \end{equation}
 \noindent where $\overline{Q_{t/h}}$ is the average hourly BC emission rate over all detected plumes, $f_{\mathrm{flare}}$ the frequency of flaring activity out of all tile acquisitions, and $f_{\mathrm{plume}}$ the frequency of BC plume detections among active flaring acquisitions.

\subsection{Supplementary figures}\label{appendix_figs}

\renewcommand{\thefigure}{S\arabic{figure}}
\setcounter{figure}{0}

 \begin{figure}[h!]
    \centering
    \includegraphics[width=13cm]{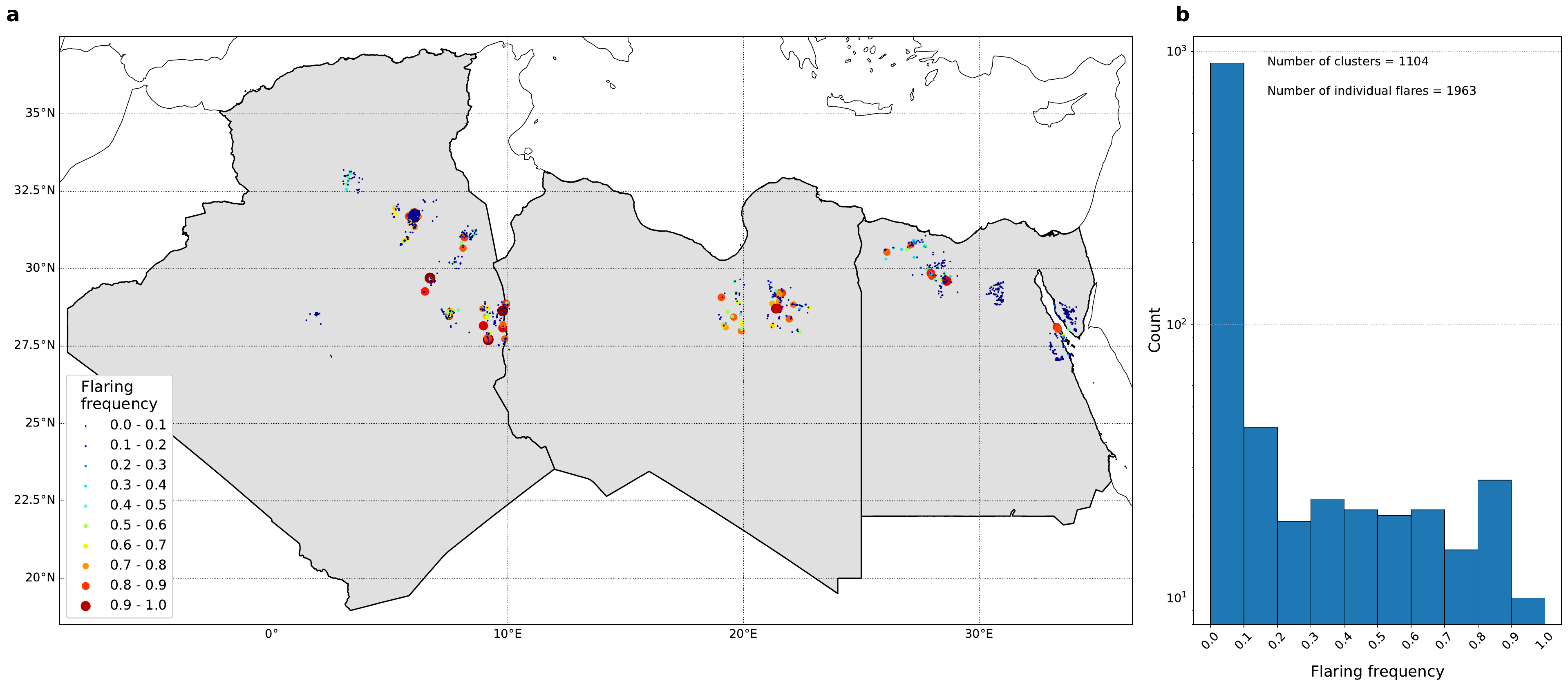}
    \caption{\textbf{Identifying flares across North Africa.} (a) Spatial distribution of the clusters of flares across North Africa. The size and colour of the dots depends on the annual flaring frequency of each location. (b) Distribution of flaring frequencies.}
    \label{Fig_S1}
 \end{figure}

 \begin{figure}[h!]
    \centering
    \includegraphics[width=9cm]{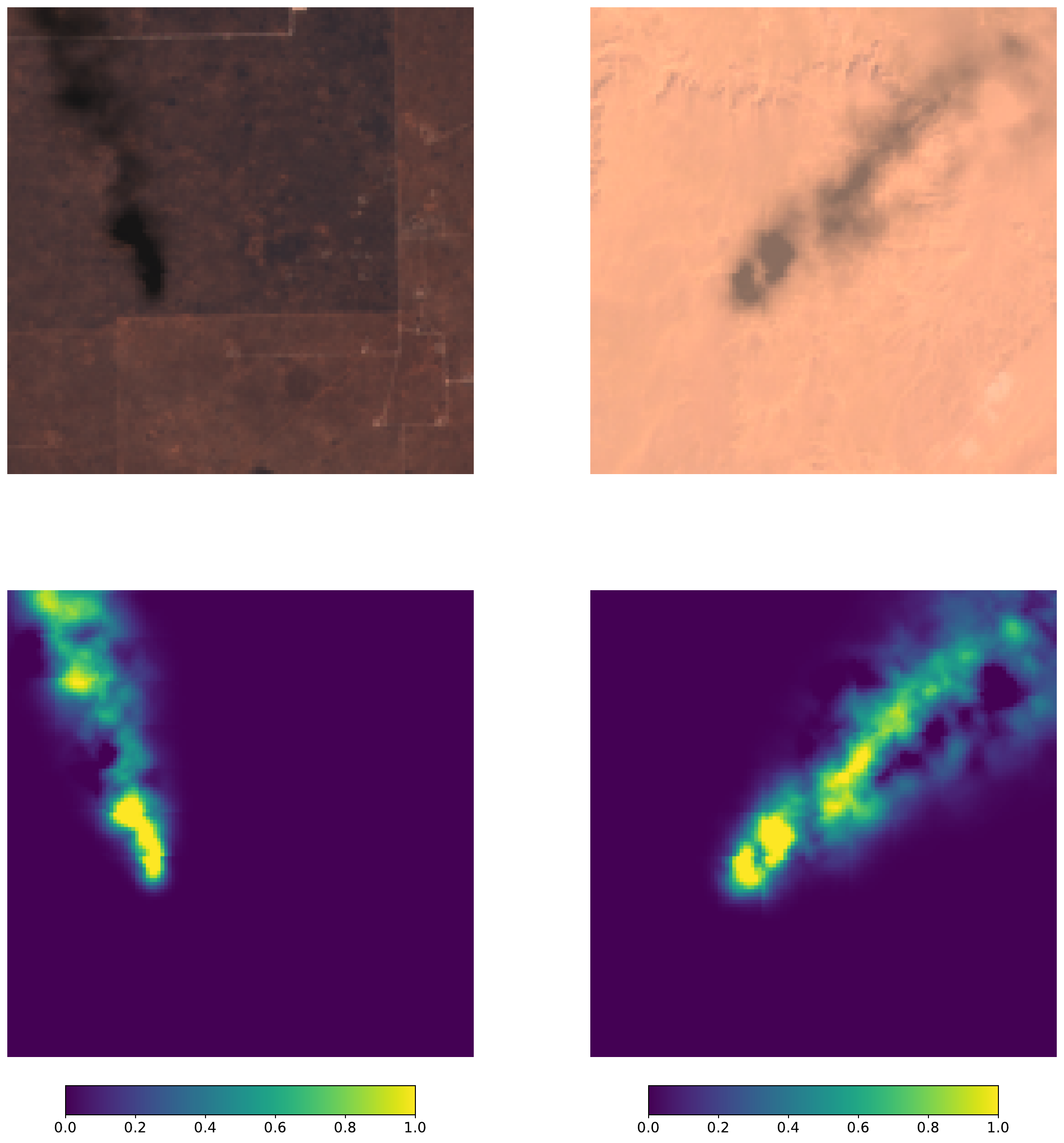}
    \caption{Example of synthetic BC plume integration in Sentinel-2 RGB scenes (top panels). The corresponding plume simulations are shown on the bottom panels.}
    \label{Fig_S2}
 \end{figure}

 \begin{figure}[h!]
    \centering
    \includegraphics[width=8cm]{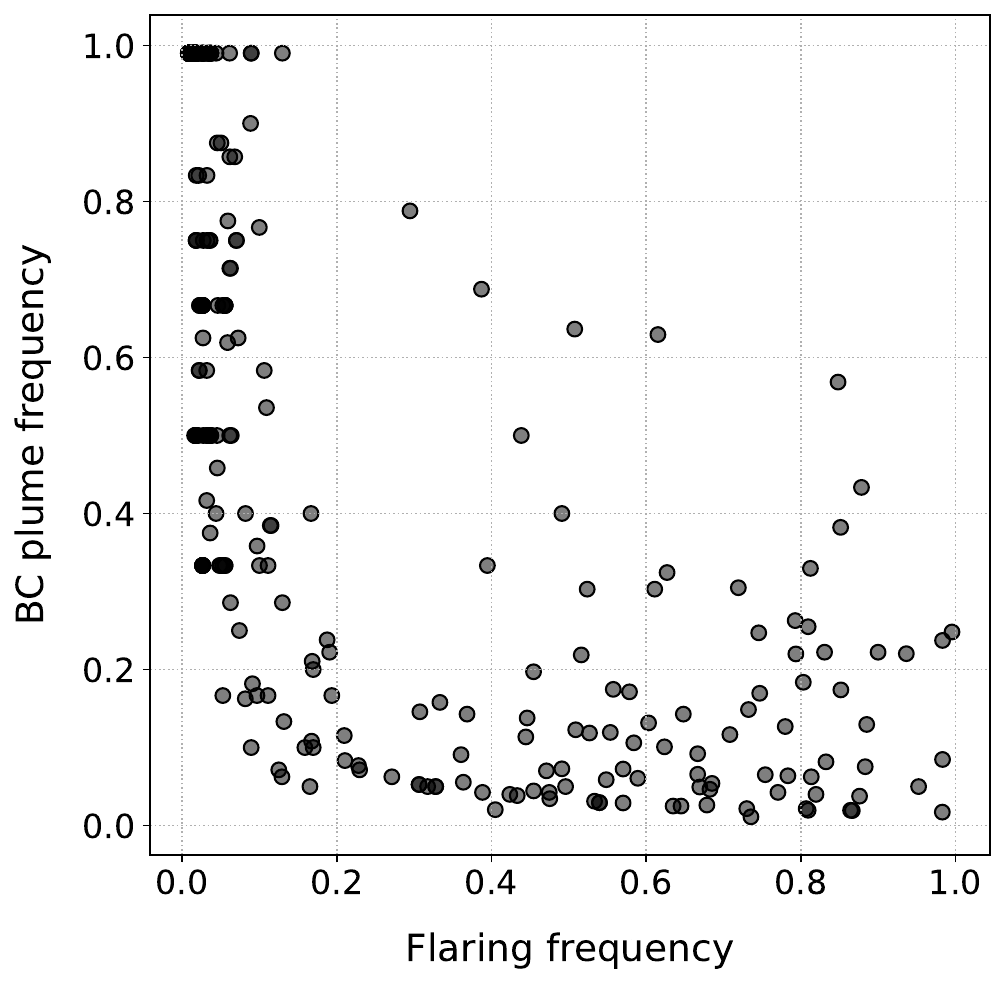}
    \caption{Map of BC emission rates across North Africa. The size and colour of the dots depends on the annual flaring frequency of each location.}
    \label{Fig_S3}
 \end{figure}

 \begin{figure}[h!]
    \centering
    \includegraphics[width=14cm]{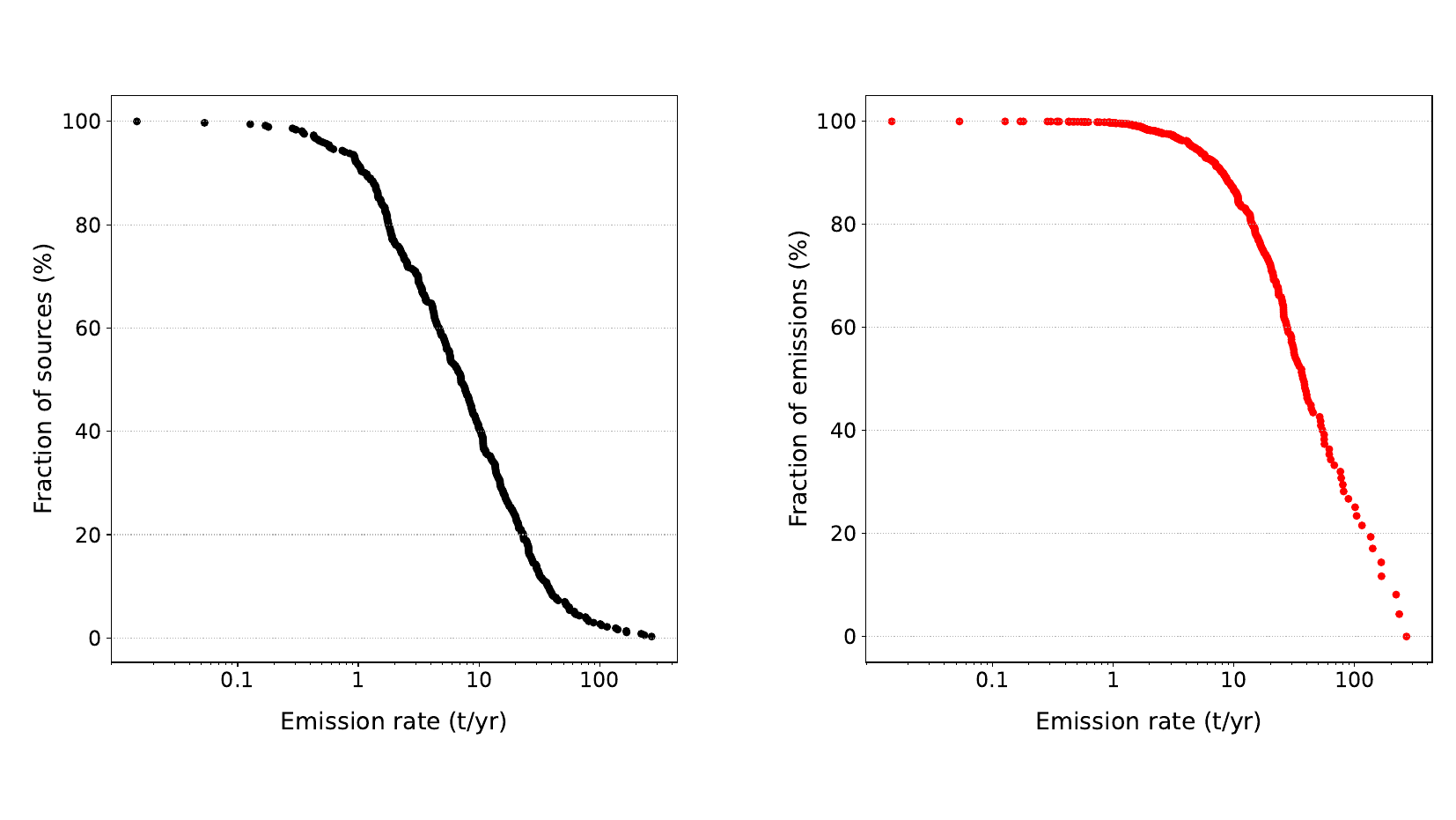}
    \caption{Cumulative frequency distributions of BC source rates (left) and of total BC emissions (right) for the 372 flare clusters with detected BC plumes.}
    \label{Fig_S4}
 \end{figure}

\end{document}